\documentclass{article}
\pdfoutput=1
\usepackage{arxiv}

\usepackage[utf8]{inputenc} 
\usepackage[T1]{fontenc}    
\usepackage{hyperref}       
\usepackage{url}            
\usepackage{booktabs}       
\usepackage{amsfonts}       
\usepackage{amsmath}
\newcommand\numberthis{\addtocounter{equation}{1}\tag{\theequation}}

\usepackage{nicefrac}       
\usepackage{microtype}      
\usepackage{lipsum}
\usepackage{graphicx}
\usepackage{float}
\usepackage[natbibapa]{apacite}
\usepackage{epigraph}

\usepackage{algpseudocode,algorithm,algorithmicx}

\usepackage{graphicx}
\usepackage{subcaption}
\usepackage{float}

\newcommand{\KL}{D_\mathrm{KL}}

\begin{document}
\title{Towards a Mathematical Theory of Abstraction}

\author{
 Beren Millidge \\
  MRC Brain Networks Dynamics Unit\\
  University of Oxford\\
  \texttt{beren@millidge.name}
  }
  
  

\maketitle   

\begin{abstract}
While the utility of well-chosen abstractions for understanding and predicting the behaviour of complex systems is well appreciated, precisely what an abstraction \emph{is} has so far largely eluded mathematical formalization. In this paper, we aim to set out a mathematical theory of abstraction. We provide a precise characterisation of what an abstraction is and, perhaps more importantly, suggest how abstractions can be \emph{learnt directly from data} both for static datasets and for dynamical systems. We define an abstraction to be a small set of `summaries' of a system which can be used to answer a \emph{set of queries} about the system or its behaviour. The difference between the ground truth behaviour of the system on the queries and the behaviour of the system predicted only by the abstraction provides a measure of the `leakiness' of the abstraction which can be used as a loss function to directly learn abstractions from data. Our approach can be considered a generalization of classical statistics where we are not interested in reconstructing `the data' in full, but are instead only concerned with answering a set of arbitrary queries about the data. While highly theoretical, our results have deep implications for statistical inference and machine learning and could be used to develop explicit methods for learning precise kinds of abstractions directly from data.

\end{abstract}
\section{Introduction}

\epigraph{
    We enunciate a rather basic principle, which might be dismissed as an obvious triviality were it not for the fact that it is not recognized in any of the literature known to this writer: \textit{If any macrophenomenon is found to be reproducible, then it follows that all microscopic details that were not reproduced, must be irrelevant for understanding and predicting it. In particular, all circumstances that were not under the experimenter's control are very likely not to be reproduced, and therefore are very likely not to be relevant.}}{\textit{Macroscopic Prediction (1985)} by E.T Jaynes}

In this paper, we present the beginnings of a mathematical theory of \emph{abstraction}. While the intuitive concept of an abstraction has long been recognized, as has the centrality of well-chosen abstractions for understanding the world, the concept itself has largely eluded mathematical formalization. In this paper, by combining insights from maximum entropy theories of statistical mechanics \citep{jaynes2003probability}, methods of statistical and Bayesian inference \citep{casella2021statistical,wainwright2008graphical,knill1996perception}, and recent advances in machine learning \citep{bishop2006pattern,goodfellow2016deep}, we aim to provide a mathematically precise definition of abstraction, understand how our definitions relate to classical methods in statistics and machine learning, and offer suggestions for how these definitions and insights can aid the development of powerful algorithms for learning abstractions from data.

Before we dive into the formalism, however, we wish to provide an intuitive and accessible description of our ideas. To begin with, we consider the `folk notion' of an abstraction. Several aspects of this slippery idea are immediately apparent. An abstraction involves some kind of summary of a more complex reality. It involves ignoring or throwing away certain kinds of information, and somehow preserving only those aspects which are most relevant to the problem at hand. Physics furnishes many examples of extremely clear and precisely specified abstractions, so in this paper we often turn to these as exemplar cases

For instance, let us consider the classical physical abstraction of a uniform mass rolling along a frictionless plane. What is the fundamental nature of such an abstraction? The first, and most obvious, property is that the objects described in the abstraction are `ideal' in some sense. The mass is completely uniform, the plane is completely frictionless and also implicitly entirely flat and smooth. These properties are not true of any masses and planes in the `real world' -- but we take these mundane real world details to be irrelevant. The second property is that due to their idealness and simplicity, the objects in the abstraction possess are vastly simpler to represent mathematically. We can usually define the mass using only several scalar variables -- its mass, and its position, velocity, and acceleration.  The frictionless plane is even simpler to represent -- typically just its angle. By contrast, a `real' mass and a `real' plane would consist of many trillions of atoms, each with their own mass and position and velocity. The combined interactions of all these vast numbers of variables would produce the actual behaviour of the system. From this, we can get to our first fundamental property of an abstraction -- an abstraction represents a \emph{compression}, often a dramatic one. In the case of the abstract mass and plane we have gone from a practically infinite amount of variables, to a finite and very small number. 

The second key property is that abstractions ignore irrelevant information. Encoded into the abstraction of the frictionless plane is the idea that the positions and velocities of the atoms comprising the plane somehow do not matter to the things we are trying to represent or use the abstraction for. This idea, while seemingly obvious, contains the two insights required to make our definition. The first insight is that abstractions are fundamentally predictive and that they are \emph{for a purpose}. An abstraction only focuses on, or is specific to, certain facets of the system, certain properties that we care about or certain questions that we wish to ask. In the case of the mass and plane, these questions are probably things about motion of the mass -- the speed of motion, the acceleration, perhaps the vector direction, and so forth. Specifically, we do not care about aspects of the real situation like the velocity of a random particle in the plane, the nature of the atmosphere surrounding the mass (if any), the precise shape of the mass, the temperature of the mass, the precise time of day this experiment is occurring, whether the experiment is occurring in the north or the south hemisphere of the earth, or in outer space. By choosing only to care about questions which do not heavily depend on this `irrelevant' information, we can create an abstraction which does not represent it. Conversely, the quality and utility of an abstraction is entirely dependent on the questions we wish to use it to reason about. The goodness of any abstraction is relative to the queries one makes of it. It is important to note that an abstraction is not necessarily just a coarse-graining of a system \footnote{Although the converse holds: a coarse graining is, of course, an abstraction}. For instance, consider again the idea of a frictionless plane. This is not simply a coarse-grained version of a plane with friction, but rather a model which ignores some aspects of the behaviour of the real system -- i.e. friction. While incorrect as a model of the real system, this nevertheless can be a good abstraction if the queries we are interested in are not significantly affected by the presence or absence of friction. 

As a consequence, abstractions are almost never perfect. By their very nature, they throw away information. That information almost always has effects, even if sometimes they are minuscule. In computer science, this fact is known by the aphorism that `all abstractions leak' -- that some detail of the underlying structure which is abstracted away always makes it through to result in actual behaviour subtly different to that predicted solely based on the abstraction. We will use this idea to formalize the `goodness' of an abstraction. Namely, that an abstraction is good to the degree to which it does not `leak' for the kinds of queries or questions that we are interested in. We operationalize the leakiness of an abstraction as the difference between the behaviour of the system (as evaluated by our queries) expected given only the abstraction, and the real behaviour.

This, then, leads us straight to the final obstacle that must be surpassed before we can make our definition. What is: `\emph{The behaviour of the system given only the abstraction}'? Here we turn to the principle of maximum entropy \citep{jaynes1957information,jaynes1988relation,jaynes2003probability} for a solution. The principle of maximum entropy is a variational principle which, originally introduces by Gibbs \citep{gibbs1879equilibrium}  and developed and formalized by Jaynes \citep{jaynes1957information,jaynes1988relation,jaynes1989clearing} underpins much of the field of statistical mechanics, and is applicable much more broadly. The key insight of this principle is that, given a system with unknown state, and some knowledge about it, you should infer the state to be the one that \emph{maximizes your ignorance (entropy), constrained by your knowledge}. For instance, given a box of gas, what should you infer the distribution of gas molecules inside the box to be? The principle of maximum entropy states that you should infer the distribution to be uniform, since it represents the maximum level of uncertainty you could have about the distribution. If you then know the average energy of the gas, you can consider this knowledge as a constraint on the ultimate distribution and instead find the maximum entropy distribution which satisfies this constraint which, in this case, is the well-known Maxwell-Boltzmann distribution.

Here, we argue that the principle of maximum entropy provides precisely the right description of what it means to throw away information irrelevant to the abstraction -- namely that such a system is the maximum entropy system where the constraints are the abstraction state. Such a system would represent our best guess at the true system state given our knowledge only of the abstraction state. If we could compare the behaviour of this maximum entropy system and the real system on a set of queries, then this would provide a direct way of measuring the leakiness of the abstraction on the questions we care about. If we have a way to measure the leakiness of an abstraction, we can turn this around and parametrize the abstraction function and then \emph{learn} those parameters to find the best abstractions possible, given data or samples from the system. This intuition is what we aim to formalize and discuss for the rest of the paper.

Finally, we turn to the quote by Jaynes at the beginning \citep{jaynes1985macroscopic} which gets to the heart of why this definition of abstraction might work, and why it may capture the important aspects of our intuitive understanding of abstraction. The key message is encapsulated in the quote at the top of this section -- that if a macroscopic phenomena can exist reproducibly, then the vast majority of microscopic interactions cannot meaningfully affect the macroscopic phenomena, since the microphenomena cannot all be controlled by the experimenter, and are therefore almost certainly not reproduced, while the macrophenomena is reproduced. From the perspective of the macrophenomena, all these microscopic interactions, and the information they contain, is irrelevant, and contains no predictive power over and above knowledge of the macrophenomena itself.
The intuition behind our definition of abstraction is precisely the reverse of this insight. If the vast majority of the microscopic interactions are irrelevant, and contain no useful information about the evolution of the system not given by the macrophenomena, then we can say that the macrophenomena provides a good \emph{abstraction} of the microscopic details involved in the overall functioning of the system. Notably, this intuition covers exactly the same intuitive bases for our normal use of the word abstraction -- the macrophenomena \emph{discards} irrelevant microscopic information while retaining just enough information to maintain itself as a reproducible phenomenon.

Additionally, we then take a more speculative turn and consider the question of why we should expect there to be useful abstractions of systems at all. We argue, albeit on an intuitive basis, that in general for many systems there should exist useful or `natural' abstractions which are useful for many different queries (i.e. generalize across queries) and which represent important `objective' abstract facts about the system in question. These `natural facts' or macrophenomena arise due to the nature of many stochastic interacting processes which tend to preserve certain kinds of information while losing others. This leads to the convergence of distributions over such systems to a few well known limiting aggregate distributions which are well studied in statistics and can be described faithfully with just a few parameters. These parameters can then be interpreted as abstractions, or macrophenomena, of the system.

\subsection{The Maximum Entropy Principle}
The maximum entropy principle originated in statistical physics, where it was first invented and applied by Boltzmann and Gibbs, and then further developed and exposited by Jaynes in a number of works \citep{jaynes1957information,jaynes2003probability} where he aims to formulate and understand the basic principles of statistical mechanics purely in terms of this maximum entropy approach, as well as apply its precepts more widely. The maximum entropy principle states, essentially, that given a system where there are many unknown microscopic degrees of freedom, but yet we know several macroscopic variables which constrain these degrees of freedom, we can predict the distribution of the system, or our measurements of the system in terms of the distribution which matches our knowledge while otherwise maintaining the highest possible entropy. In mathematical terms, the maximum entropy principle states that the distribution of the system will be that which maximises its own entropy while simultaneously satisfying any constraints imposed by our knowledge of the system, which usually take the form of macrophenomonal averages over system states. From a thermodynamic perspective, the maximum entropy principle provides a heuristic but stronger version of the second law of thermodynamics, stating not just that entropy will increase, but that it will tend to increase to the \emph{maximum} possible allowed by the intrinsic constraints of the system.

Intuitively, the maximum entropy principle can be justified by the fact that it presupposes the least knowledge about the system -- we explicitly encode the fact that we are maximally ignorant about everything other than what we explicitly know. The maximum entropy principle thus generalizes Laplace's principle of indifference as a special case where we in fact know nothing at all about the distribution except its support and so derive a uniform distribution. Mathematically, the maximum entropy principle prescribes a straightforwardly applicable apparatus for computing maximum entropy distributions given a certain set of constraints using the simple principle of constrained optimization with Lagrange multipliers. For instance, consider the original thermodynamical problem where we wish to describe the distribution of states of the system $p(x_i)$ where $x_i$ is the i'th state, given an average energy $\bar{E} = \sum_i p(x_i) E_i$. The maximum entropy principle states that we can solve this by solving the following constrained optimization problem,
\begin{align*}
    p^*(x) &=\underset{p(x)}{argmax} \mathcal{L} \\
    \mathcal{L} &=
    -\int dx \, p(x) \ln p(x) - \lambda_0 \big( \int \, dx p(x) - 1) - \lambda_0 (\int dx \, p(x) E(x) - \bar{E}) \numberthis
\end{align*}
Where the entropy $\mathbf{H}[p(x)] = -\int dx \, p(x) \ln p(x)$ is maximized while the first constraint with multiplier $\lambda_0$ encodes the normalization constraint that the probability distribution over the states must sum to 1, and the second constraint encodes the fact that the average energy of the system $\sum_i p(x) E(x)$ must equal the known average energy $\bar{E}$. To solve, we take the derivatives and set them equal to 0,
\begin{align*}
    \frac{\partial \mathcal{L}}{\partial p(x)} &= 1 + \ln p(x) - \lambda_0 - \lambda_1 E(x) =0 \\
    &\implies p^*(x) = e^{-(\lambda_0 - 1)}e^{-\lambda_1 E(x)} \numberthis
\end{align*}
Then, plugging this into the normalization constraint, we can derive,
\begin{align*}
    \int dx \, e^{-(\lambda_0 -1)}e^{-\lambda_1 E(x)} &= 1 \\
    e^{-(\lambda_0 - 1)} \int dx \, e^{-\lambda_1 E(x)} &= 1 \\
    e^{\lambda_0 -1} &= \int dx \, e^{-\lambda_1 E(x)} \\
     \lambda_0 - 1 &= \ln  \int dx \, e^{-\lambda_1 E(x)} = \ln Z \numberthis
\end{align*}
where $Z = \int dx \, e^{-\lambda_1 E(x)}$ is the partition function or normalizing constant. With this definition, we can thus write,
\begin{align*}
    p^*(x) = \frac{1}{Z}e^{-\lambda_1 E(x)} \numberthis
\end{align*}
If we do some more algebra, and apply some thermodynamic knowledge (see \citet{jaynes1957information} for details), we can derive that $\lambda_1 \propto \frac{1}{T}$ where $T$ is the temperature, thus deriving the Boltzmann distribution. In Appendix A, to showcase the power of the framework, we also derive the Gaussian (or normal) distribution from the maximum entropy framework under the constraint that we only know the variance of the distribution $\int dx \, p(x) (x - \mu)^2 = \sigma^2$.

In general, all maximum entropy problems share the same mathematical apparatus and take a common form, differing only in the details of the constraints. Specifically, for a general set of constraints $\mathbb{\lambda} . \mathbb{C} = \sum_i \lambda_i \big( f(x) - c_i \big)$, we can express the maximum entropy objective as,
\begin{align*}
    \mathcal{L} = -\int dx p(x_i) \ln p(x_i) - \lambda_0 \big( \int dx p(x) - 1 \big) - \sum_i \lambda_i \big( f(x) - c_i \big) \numberthis
\end{align*}
with the solution,
\begin{align*}
    \frac{\partial \mathcal{L}}{\partial p(x)} = 0 \implies p^*(x) = e^{-(\lambda_0 - 1)}e^{-\sum_i \lambda_i f(x)} \numberthis
\end{align*}
where $\lambda_0 - 1 = \ln Z = \ln \int dx \, e^{-\sum_i \lambda_i f(x)}$ relates to the log partition function or, in probabilistic terms, the normalizing constant for the density since it arises from the normalization constraint. The other lagrange multipliers $\{ \lambda_1 \dots \lambda_N \}$ can often be solved for via algebraic manipulations over the expressions for the constraints, which will ultimately give an analytical expression for the maximum entropy distribution. Furthermore, we always know that the maximum entropy objective is a minimum by taking the second derivative, by which we find that $\frac{\partial^2 \mathcal{L}}{\partial p(x)^2} = \frac{1}{p(x)}$ which is always positive.
    
\section{Formulation}

While previously we have given an intuitive description of the key ideas behind our definition, here we provide the mathematical formalism to make these ideas precise. To recap concisely, an abstraction is a small set of 'summary statistics' which provide sufficient information to answer queries about a system such that, for the queries we wish to ask, the answers to the queries of the particular system are the same as the answers given by the maximum entropy system corresponding to the constraint that the abstract variables are set to the values they hold. 

Mathematically, let us define a system as possessing variables $x \in \mathbb{R}^\mathbb{X}$ where $\mathbb{X}$ is the dimensionality of the state vector, which we assume to be very large in general. If we are uncertain about the system, we can define a probability distribution $p(x)$ over the variables of the system. Secondly, we define an \emph{abstraction} $a \in \mathbb{R}^\mathbb{A}$ to be a vector of abstraction variables where $\mathbb{A} << \mathbb{X}$ is the dimensionality of the abstraction vector, and is usually assumed to be much smaller than the dimensionality of the system. Intuitively, one can think of $a$ as a kind of  `summary statistic' for the system in a way we shall make precise shortly. For now, we assume that we have a given $a$, and we will discuss learning $a$ from data later. We define the maximum entropy function $m$ such that  $m(a) = \tilde{p}(x | a)$ is a function that maps a given abstraction variable to the maximum entropy distribution over the system's variables, given the constraints of the abstraction variables following a particular distribution, where we write $\tilde{p}$ to denote the fact that this is a maximum entropy distribution.  Formally, this can be written as,

\begin{align*}
\label{maxent_variational_problem}
m(a) = \tilde{p}(x | a) = \underset{p(x | a)}{argmax} \, \mathbf{H}[p(x | a)] - \lambda_0 \big( \int p(x |a)dx - 1 \big) \numberthis
\end{align*}
Where $\lambda_0$ is the lagrange multiplier which ensures that the probabilities of each state sum to one. We also possess a set of \emph{query functions} $\mathbb{Q} = \{ Q_1, Q_2 \dots Q_N \}$; $ Q: \mathbb{X} \rightarrow \mathcal{Q}$, where $\mathcal{Q}$ is the output space of the query function, which may be different from the state space, and different between different query functions. For generality, we assume that the query functions can themselves be probabilistic and and therefore map from a distribution over $\mathbb{X}$ to a distribution over $\mathcal{Q}$: $p(q | x) = Q(p(x))$ given a distribution over system states $p(x)$. From these definitions, we define the `goodness', or `loss' $\mathcal{L}_Q(a,x)$ of a given abstraction, for a given query and a given system, to be the divergence between the query distribution given the true system, and the query distribution taken from the maximum entropy system inferred from the constraints. We write this as,
\begin{align*}
\label{Abstraction_objective}
    \mathcal{L}_Q(x,a) = \mathcal{D}\big[ Q(p(x)) || Q(m(a)) \big] \numberthis
\end{align*}
In general, we take the divergence function $\mathcal{D}$ to be the KL divergence $\mathcal{D}_{KL}[p(x)||q(x)] = \int dx p(x) \log \frac{p(x)}{q(x)}$, although other divergences are possible. If we have a set of query functions, we can define the overall goodness of the abstraction to be simply the average loss over all queries,
\begin{align*}
\label{query_average_obj}
    \mathcal{L}(x,a) = \sum_{Q_i \in \mathbb{Q}} p(Q_i) \mathcal{L}_{Q_i}(a,x) \numberthis
\end{align*}
This equation provides a quantitative measure of the goodness of fit of a particular set of abstractions given the set of queries one might want to make of that abstraction. In computer science terminology, Equation \ref{Abstraction_objective} measures the `leakiness' of an abstraction. If $\mathcal{L} = 0$ then knowing the abstraction allows us perfectly to predict all queries without needing to know any more information about the original system than is contained within the abstraction. If $\mathcal{L} > 0$, then there is some information about the system, which is relevant to our queries, which is not captured by the abstraction -- the abstraction `leaks'.

We define a \emph{perfect abstraction} for given queryset and system to be the abstraction $a$ for which $\mathcal{L}(a,x) = 0$. If our queryset is the set of all possible queries, then we can define a \emph{universally perfect abstraction}, for which the abstraction loss is 0 for all possible queries. A universally perfect abstraction can only exist when $p(x) = \tilde{p}(x | a)$ -- i.e. that the real system distribution is already a maximum entropy distribution which can be specified with abstraction constraints $a$ which would simply be the parameters of the maximum entropy distribution. In statistics, universally perfect abstractions of a distribution are called sufficient statistics, since they contain all the necessary information to precisely recreate the distribution, and thus answer any query about it. By this argument, it is clear that sufficient statistics can only exist for maximum entropy distributions, thus providing an intuitive grounding for the Pitman-Koopman-Darmois theorem \citep{pitman1936sufficient,koopman1936distributions} which states that finite-size sufficient statistics only exist for exponential distributions -- i.e. maximum entropy distributions. Conversely, there are some (trivial) query sets, for which all abstractions are perfect. This can occur with any query function which simply ignores its input, and returns the same output for every input. All abstractions under this (trivial) query are perfect. 

Given a perfect abstraction for a given query, we can directly show that this means that it contains all the information about the system important in answering the query. In information-theoretic terms, this means we can show that the mutual information between the final query distribution and the abstract summary is the same as the mutual information between the query distribution and the actual state, thus implying that all the information relevant to the query is captured by the abstraction
\begin{align*}
\label{infotheory_proof}
    \mathcal{L}_Q(a,x) = 0 &\implies \KL \big[ Q(p(x)) || Q(m(a)) \big] = 0 \\
    &\implies Q(p(x)) = Q(m(a)) \\
    \\
    \mathcal{I}[q ; x] &= \KL \big[ p(q,x)||p(q)p(x) \big] \\
    &= \KL \big[Q(p(x)) || p(q) \big] \\
    &= \KL \big[  Q(m(a)) || p(q) \big] \\
    &= \KL \big[ p(q | a) || p(q) \big] \\
    &= \mathcal{I}[q;a] \numberthis
\end{align*}
If we then generalize slightly from being given a specific abstraction vector, to instead being given a distribution $p(a)$ over the abstraction vector, we find that the information contained within the abstraction with the queries is greater than that just within the system, with the difference coming from the entropy of the abstraction distribution.
\begin{align*}
    \mathcal{L}_Q(x, p(a)) = 0 &\implies \mathbb{E}_{p(a)} \KL \big[ Q(x) || Q(m(a))p(a) \big] = 0 \\
    &\implies Q(x) = Q(m(a))p(a) \\
    \mathcal{I}[q ; x] &= \mathbb{E}_{p(a)} \KL \big[ p(q,x)||p(q)p(x) \big] \\
    &= \mathbb{E}_{p(a)} \KL [Q(m(a))p(a) || p(q)] \\
    &= \mathcal{I}[q;a] - \mathbf{H}[p(a)] \numberthis
\end{align*}
If the abstraction distribution becomes a delta distribution $p(a) = \delta(a -\bar{a})$ then $\mathbf{H}[p(a)] = 0$ and so we regain the original equality.

\begin{figure}
    \centering
    \includegraphics[scale=0.25]{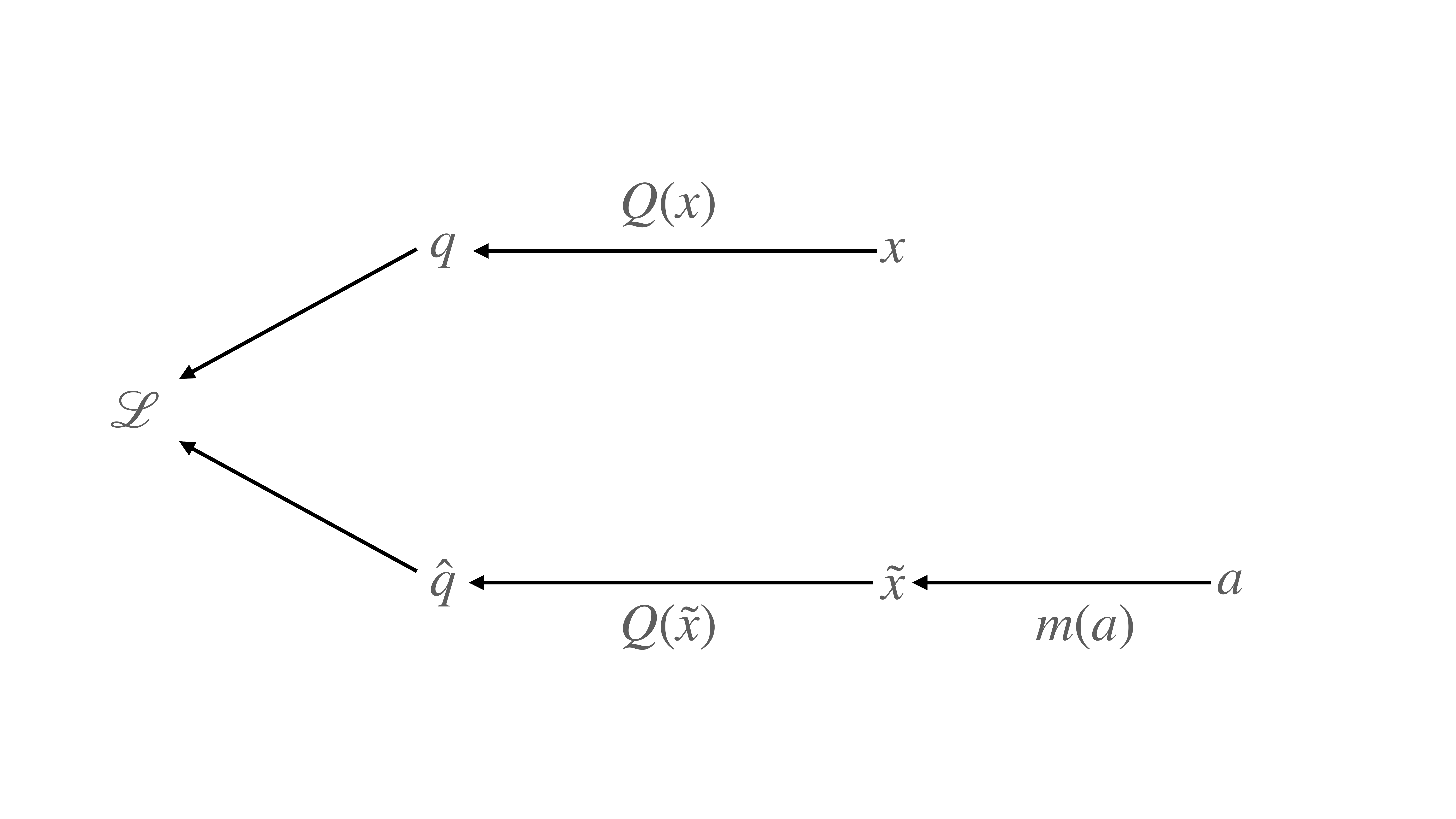}
    \caption{The logical flow of the abstraction objective. We compare the query function evaluated on the real system $q = Q(x)$ with that of the query set evaluated on the maximum entropy distribution $\hat{q} = Q(\tilde{x})$ which is generated from the abstraction $\tilde{x} = m(a)$. The idea is that the difference between the `true' queryset evaluated on the system and that generated through the maximum entropy process based only on the abstraction is a good measure of the `leakiness' of the abstraction, since if this difference is minor, then this abstraction provides a good representation of the aspects of the system which we care about (as measured by the queries).}
    \label{fig:my_label}
\end{figure}

Figure 1 shows a graph of the pattern of logic required. A good abstraction is one which minimizes the divergence over the query-set between the real system, and the maximum entropy system given the constraints imposed by the abstraction.

\subsection{Learning abstractions}

While thus far, we have only provided a definition of an abstraction, and a means to judge the goodness of one, an important question we still wish to answer is whether we can \emph{learn abstractions from data}. Given that we possess a computable objective function which measures goodness of an abstraction, this is straightforward in principle. We suppose we are given a dataset of $i.i.d$ instances of the system $D = \{ x_1 \dots x_N \} \sim p(x)$ drawn from the distribution over system states from which we can learn the abstraction vector which minimizes our abstraction objective.

First, for conceptual clarity, we must distinguish between two different possible types of abstraction variables. Firstly, we have fixed, or static abstraction variables which do not depend directly on specific data items but only on the dataset as a whole. These are invariants of the dataset, and correspond to the classical notion of parameters in statistical models. We denote these as $a$ and we often optimize a distribution over them $p(a;D) = p(a)$ where we use the $;D$ when it is necessary to make explicit that these depend on the dataset as a whole, but are not a direct function of a specific dataset. The second kind of abstraction variables that are possible are those that depend \emph{directly} on specific datapoints. In machine learning models these correspond to the latent variables which are output by an encoder network but which change depending on the specific datapoint considered and are used to model that specific datapoint. We denote these abstractions as $\bar{a}$ and their distribution as $p(\bar{a} | x; \theta)$, which is the output of an abstraction function $\bar{a}(x;\theta) = p(\bar{a} | x)$ with a set of parameters $\theta$ which are learnt over the whole dataset and then applied to each datapoint. For instance $p(\bar{a} | x; \theta)$ could represent an encoder network in a variational autoencoder \citep{kingma2013auto} where the parameters $\theta$ are the weights of the neural network (and, since they are not direct functions of the datapoint, are in $a$). For now, we will generally only consider the classical statistical parameters $a$ as abstractions, and will return to this notion of latent variables encoding datapoints when we discuss the relationship to machine learning.

Combining these two sets of abstraction variables, we can write the joint objective to be optimized, for a given datapoint $x$ as,
\begin{align*}
\label{full_objective}
\mathcal{L}_Q(x, p(a), \theta) = \mathbb{E}_{p(a)}\KL \big[ Q(x) || Q(m([a,\bar{a}]))p(a) p(\bar{a} | x;\theta) \big] \numberthis 
\end{align*}
Then, we can simply learn the necessary abstraction parameters $p(a), \theta$ as an optimization process over the full dataset -- for instance using gradient descent,
\begin{align*}
\label{gradient_descent}
    \dot{\theta} = - \mathbb{E}_{p(D)}\big[ \frac{\partial \mathcal{L}_Q(x,p(a),\theta)}{\partial \theta} \big] \numberthis
\end{align*}
And similarly for $p(a)$. If we wish, we could optimize both $a$ and $\bar{a}$ simultaneously by interchanging iterations of optimizations of one variable while holding the other fixed. In the language of Bayesian inference, this corresponds to a factorization assumption between $a$ and $\bar{a}$ where inference is performed through the Expectation-Maximization (EM) algorithm \citep{dempster1977maximum}. This objective is computable as long as the queries are computable and the solution of the maximum entropy variational problem to infer $\tilde{p}(x | a)$ is tractable. As such, the gradients of the objectives required for Equation \ref{gradient_descent} can be computed by automatic differentiation software \citep{paszke2017automatic,griewank1989automatic}. An important note is that if we wish to use the abstractions for multiple queries in a query set, we should update the learning rule to update with the average abstraction loss over the query set. We will discuss the implications of this for machine learning architectures later.
\begin{align*}
    \dot{\theta} = - \sum_{Q_i \in \mathcal{Q}} p(Q_i) \mathbb{E}_{p(D)}\big[ \frac{\partial \mathcal{L}_Q(x,p(\bar{a}),\theta)}{\partial \theta} \big] \numberthis
\end{align*}

While this approach will be able to infer the best abstraction vector given the state, it can only infer abstraction vectors of a given dimensionality. However, for many systems, the correct dimensionality of the abstractions is not known a-priori. Thus, more general methods which can infer both the optimal $a$ and the optimal dimensionality of $\mathbb{A}$ would ultimately be ideal. This problem is very close to that of model selection encountered in the statistics literature, and may be solved in a very similar manner. For instance, if we consider the dimensionality of $\mathbb{A}$ as a model, then we could fit separate functions $p(\bar{a} | x; \theta)$ for each dimensionality, then compare the ultimate abstraction loss obtained, and ultimately compare the models through classical model selection techniques such as Bayes Factors \citep{sakamoto1986akaike,stephan2009bayesian,chipman2001practical}. 

One key computational difficulty with this scheme is the necessity to infer the maximum entropy distribution before applying the query distribution, since the query distribution takes in the system distribution $p(x)$ as input. While in many cases inferring the maximum entropy distribution of the state is easy since, for well-known constraints, it follows a natural and mathematically tractable maximum entropy distribution, such as the Boltzmann, or Gaussian distributions, in many other cases, with more complex constraints, the solution to the variational equation in Eq \ref{maxent_variational_problem} is a challenging computational problem without a neat solution. In these cases, there may be difficulty directly computing, and hence optimizing this loss function to be able to learn good abstractions.

The ideal scenario would be if we were able to define or learn a composite query function which directly maps from the abstract state to the query distribution -- $\tilde{Q}: \mathbb{A} \rightarrow \mathcal{Q}$. By inspecting Figure 1, we can see that this composite query function is simply the composition of the maximum entropy function $m$, which maps the abstraction to its maximum entropy distribution and the original query function. Thus we have that $\tilde{Q} = Q \circ m$. If this composite function $\tilde{Q}$ could be learnt, approximated, or computed straightforwardly, it would greatly improve the computational efficiency of this scheme by circumventing the necessity to solve the maximum entropy variational problem on each step. Additionally, it has the benefit of conceptual clarity. In many cases, the whole point of abstraction is to not need to refer to the original system anymore, and instead run your queries directly on the abstract representation. This is precisely what $\tilde{Q}$ enables. Discovering whether this scheme can be reorganized, so that $\tilde{Q}$ becomes the central computational unit is a very important area for future work, since it would bring this scheme much closer to the intuitive use of abstractions, where the queries are often defined only in terms of the abstract variables themselves and not upon the entire system state. 

If we allow for a distinction between phases of `learning' and phases of `inference', then it is quite straightforward to imagine an approximate $\tilde{Q}(a; \phi)$, with parameters $\phi$ being learned during a learning phase where the abstractions $a$ are also learnt by optimizing Eq \ref{Abstraction_objective}. The loss function for $\tilde{Q}$ would be extremely simple -- just the difference or divergence between the prediction given by $\tilde{Q}(a;\phi)$ and the actual query output by going through the maximum entropy system -- $\mathcal{L}_{\tilde{Q}} = \mathcal{D} \big[ \tilde{Q}(a;\phi)||Q(m(a))) \big]$. Then, once $\tilde{Q}$ has been learned, it could be used during an inference phase to directly infer query outputs purely from the abstractions without any reference to the underlying system.

\subsection{Dynamical Abstractions and Maximum Calibre}

It is fairly straightforward to extend this principle to the inferring abstractions of dynamical systems, which is more often the case that obtains in reality. Here, we do not wish to simply learn a static abstraction to handle static queries from a static system. Instead, we wish to be able to answer queries over time from a dynamically changing system. To handle this in the maximum generality, we therefore need to define dynamically changing abstractions, or rather an abstract dynamical system, which tracks the relevant behaviour of the real system, while throwing away all information irrelevant to the queries. This abstract dynamical system maintains a low-dimensional set of abstract variables $a_t \in \mathbb{A}$ and updates them over time in accordance with the abstract differential equation $\dot{a} = f_a(a_t)$ where $f_a$ is the abstract update function. To specify such an abstract system, in direct analogy with the low-level system, we need only specify the initial abstraction $a_0$ and the abstract update function $f_a$. Similarly, we define our particular system in terms of differential equations $\dot{x} = f(x)$ with a dynamics function $f$. We typically want our queries to remain constant over time, but can also use queries which change over time $Q(x,t)$. The key change comes in our definition of the maximum entropy system, given the constraints encoded by the abstractions. Here, instead of using the principle of maximum entropy, we instead use the principle of \emph{maximum calibre} \citep{jaynes1980minimum,dixit2018perspective}, an extension of maximum entropy, which instead considers maximizing the path integral of entropy over a dynamical trajectory, given dynamical constraints, rather than simply minimizing the maximum of the entropy at each particular point. If we denote a trajectory of variables using square brackets $[x_t] = [x_1 \dots x_T]$, then we can define the maximum calibre path as,
\begin{align*}
    m([a_t]) = \tilde{p}([x_t] | [a_t]) = \int dt \, \underset{p([x_t] | [a_t])}{argmax} \, \mathbf{H}[p([x_t] | [a_t)]] - \lambda_0 \big( \int p([x_t] |[a_t]) dx - [1] \big) \numberthis 
\end{align*}
Given this, then it is straightforward to extend our previous abstraction loss functions to instead become path integrals of the loss function over time,

\begin{align*}
    a^* &= \underset{a_0, f_a}{argmin} \, \mathcal{L}([a_t],[x_t]) \\
    \mathcal{L}([a_t],[x_t]) &= \int dt  \sum_{i \in \mathcal{Q}}, p(q_i) \mathcal{L}_Q(a,x,t) \\
     \mathcal{L}_Q(a,x,t) &= \mathcal{D}\big[ Q(x_t) || Q(m(a_t)) \big] \numberthis
\end{align*}

\subsubsection{Learning Dynamical Abstractions}

While the generalization to dynamical abstractions is straightforward, dynamical abstractions also give us an additional unique ability relative to static ones, namely that it is possible to define a contrastive objective that can learn them purely in the abstraction space, without reference to the original (or the maximum entropy version) system, if we have access to the correct composite function $\tilde{Q} = Q \circ m$. The key idea here is that with a dynamical system, where we possess both an abstract state and an abstract dynamics function, we can simply compare the dynamical evolution of the abstract state at every timestep with the abstract state inferred from the real system state at that timestep. For instance, suppose we consider the predicted abstract state according to the abstract dynamics function as $p(\hat{a}_t | a_{t-1}) = f_a(p(a_{t-1}))$, and additionally, we have the abstraction generated directly from the system state $x_t$ from the abstraction function $p(a_t | x_t) = a(x_t)$. Given that we have these two independent sources of information about the abstraction -- its a-priori abstract dynamics and the abstraction directly inferred from the current state of the system, we can simply minimize the divergence between these relative to the composite query function as our new abstraction objective.
\begin{align*}
    a^*, f_a^* = \underset{a, f_a}{argmin} \, \int dt \, \mathcal{D}\big[ Q(a_t)p(a_t| x_t) || Q(\hat{a}_t)p(\hat{a}_t | a_{t-1}) \big] \numberthis
\end{align*}
Importantly, this objective makes no reference to the underlying system except insofar as to infer the abstraction from it, and completely eschews the computation of the maximum-entropy system given the abstraction state, which may be a significant computational saving. This option is available in the case of dynamical abstractions, because we have multiple sources of information about the abstraction -- namely the abstraction inferred directly from the system, and the abstract state computed by the abstract dynamics function, which we can enforce consistency between. The key challenge in this case is learning the abstract `encoder' function $a(x_t)$ which maps from the system state $x_t$ to the desired abstraction of that state $a_t)$. If we possess a dataset of known pairs $(x_t, a_t$ then we can learn this encoder system simply by treating it as a regression problem.

\subsection{Relationship to statistical inference}

So far we have considered abstractions as a function of arbitrary queries $Q$. Here, we consider the abstraction learning objectives under one specific query -- the reconstruction query: $Q(p(x)) = p(x)$. This is an extreme case and is where the query is simply to precisely reconstruct the system as accurately as possible. In this case, we discover that our scheme precisely collapses into standard maximum likelihood (ML), or maximum a-posteriori (MAP)\footnote{Where the prior comes in as a prior over the abstraction distribution $p(a)$} inference as is undertaken in classical statistics. This result provides the important conceptual link for understanding how our notion of abstraction and abstraction learning relates to standard statistical methods -- namely that standard statistical methods can be considered to be learning abstractions under just the reconstruction query. That is, we can conceptualise statistical methods as aiming to learn those abstractions which will allow them to precisely characterise the entire dataset \emph{without throwing away any information}. Conversely, this allows us to view our abstraction approach as a straightforward generalization of classical statistics to the case where we are not interested in modelling the data itself, but only in modelling (lossy) \emph{functions} of the data.

To begin, we return to the original abstraction objective (Equation \ref{Abstraction_objective}) and apply it to the case where we possess a dataset of $i.i.d$ elements drawn from the distribution over the system $D = \{ x_1 \dots x_N \} \sim p(x)$. In this case, the optimal abstraction is,
\begin{align*}
    a^* &= \underset{a}{argmin} \, \mathbb{E}_{p(D)} \KL[Q(x) || Q(m(a))] \\ 
    &=  \underset{a}{argmin} \,  \mathbb{E}_{p(D)} \KL[p(x) || \tilde{p}(x | a)] \numberthis
\end{align*}
Where we have used the fact that we are assuming the reconstruction objective $Q(x) = p(x)$. Then, using the fact that dataset distribution is simply an empirical distribution over observed datapoints $p(D) = \sum_i \delta(x - x_i)$, we can see that this objective simplifies to,
\begin{align*}
    a^* &= \underset{a}{argmin} \,  \mathbb{E}_{p(D)} \KL[p(x) || \tilde{p}(x | a)] \\
    &= \underset{a}{argmin} \, -\mathbf{H}[p(D)] - \mathbb{E}_{p(D)}[\ln \tilde{p}(x | a)] \\
    &= \underset{a}{argmax} \sum_i \ln \tilde{p}(x_i | a) \numberthis
\end{align*}
Where in the final line we have used the fact that the entropy of the data distribution $\mathbf{H}[p(D)]$ is fixed (with respect to the abstraction parameters) so that it has no impact on the optimization, and the fact that the data distribution is empirical over the datapoints, so the expectation simply reduces to a sum of the probabilities of each datapoint. This objective is then obviously the standard maximum likelihood objective from statistics, as it simply aims to maximize the sum of the log likelihoods of the data given the abstraction, which would correspond simply to a statistical parameter (such as the mean or variance) in a statistical generative model. An important but subtle difference, however, to classical maximum likelihood is that here we are maximizing the likelihood of the \emph{maximum entropy distribution} $\tilde{p}(x_i | a)$. In almost all cases, this will be the same as the standard likelihood $p(x_i | a)$ in classical statistics, since almost all of the distributions commonly used in parametric statistics are, in fact, maximum entropy distributions. The reason for this is simply because maximum entropy distributions are precisely those which can be parametrized with a finite number of sufficient statistics. However, this approach to may subtly differ in the case of a nonparametric likelihood model, which can be modelled with an implicitly infinite amount of parameters.

If we generalize this to instead learn a distribution over the abstractions $p(a; \phi)$ with potentially some parameters $\phi$, then the objective we optimize will become,
\begin{align*}
\phi^* &= \underset{\phi}{argmin} \, \mathbb{E}_{p(D)} \KL[Q(x) || Q(m(a))p(a; \phi)] \\
&= \underset{\phi}{argmax} \sum_i \ln \tilde{p}(x_i | a)p(a; \phi) \numberthis
\end{align*}
Where to reach the second line, we have simply redone exactly the same steps as in the previous derivation but with our new objective. It is straightforward to see that this then corresponds to maximum a-posteriori (MAP) inference where the abstraction distribution $p(a;\phi)$ becomes identified with the Bayesian prior. If we take the abstraction distribution to be deterministic $p(a;\phi) = \delta(f(\phi) - a)$, then the MAP objective collapses back down to the ML, as expected. Effectively, what we have shown is that the learning of abstractions reduces to classical ML or MAP statistical inference using a reconstruction query or, conversely, that classical statistical inference methodologies to learn latent variables can be seen as abstraction learning techniques where the query is simply to reconstruct the data. This means that our notion of abstraction considered here can be construed as a straightforward generalization of classical statistics to the case where we are not interested in reconstructing the data, but a set of lossy, or coarse-grained, functions of the data. Alternatively, we can consider statistics, or the process of learning latent variables or parameters of models, to be implicitly learning abstractions where the only query is simply to reconstruct the data as accurately as possible.

Interestingly, it is easy to show that the reconstruction objective $Q(x) = p(x)$ is, in an information-theoretic sense, the `hardest' query to abstract in that it preserves the most possible information about $x$. This is obvious intuitively, since of course $x$ (or any bijective function of $x$) contains the most possible information about $x$. It is also straightforward to show this mathematically. If we consider the information between the system state $x$ and the query distribution under two queries, the reconstruction query $q_1 = Q_1(x) = x$ and any other query $q_2 = Q_2(x) \neq x$.
\begin{align*}
    \mathcal{I}[q_1, x] &= \mathcal{I}[x,x] = \mathbf{H}[\mathbf{X}] \\
    \mathcal{I}[q_2,x] &= \mathbf{H}[\mathbf{X}] -\mathbf{H}[\mathbf{X} | \mathbf{Q}_2] \\
    &\leq \mathbf{H}[\mathbf{X}] \\
    &\leq \mathcal{I}[q_1, x]  \numberthis
\end{align*}
Where the inequality follows because the entropy is necessarily positive. Maximum likelihood statistics also has an close relationship with the principle of information-maximization (Info-Max) which is fairly straightforward and is explored in Appendix C. We briefly discuss some potential applications of our ideas for machine learning in Appendix D.
\section{Why should good abstractions often exist?}

\epigraph{\textit{I  know  of  scarcely  anything  so  apt  to  impress  the imagination   as   the   wonderful   form   of   cosmic   order expressed by  the  ‘law  of  frequency  of  error’} [the  normal or  Gaussian  distribution].  \textit{Whenever  a  large  sample  of chaotic  elements  is  taken  in  hand  and  marshaled  in  the order   of   their   magnitude,   this   unexpected   and   most beautiful  form  of  regularity  proves  to  have  been  latent all  along.  The  law...reigns  with  serenity  and  complete self-effacement  amidst  the  wildest  confusion.  The  larger the mob and the greater the apparent anarchy, the more perfect  is  its  sway.  It  is  the  supreme  law  of  unreason.}}{\textit{Natural Inheritance (1894)} by Francis Galton}

Here, we consider the question of why there should be useful abstractions in general. What is the fundamental reason that, empirically, there often appear be low dimensional summaries that are almost perfect at answering queries about very high dimensional systems -- and which can often generalize across many disparate queries? The key element again comes from the maximum entropy principle through the form of aggregating, or limiting, distributions. It is a well known fact in statistics that complex generative processes which consist of a large number of interacting parts often result in data which are highly regular and conform to one of a simple set of known and mathematically tractable distributions.  

The most well known example of this is the Gaussian distribution which, by the Central Limit Theorem as Galton poetically describes, any process which involves the summing (or convolution) of a large number of random variables, no matter the distribution of those variables (as long as the distribution has a finite mean and variance), will converge towards. This simple fact of aggregation explains why so many processes we observe in nature so often tend to obey Gaussian statistics, even if on the surface they appear to have no clear features that would predict Gaussianity. While this effect is most well known for Gaussians, there are a wide range of such aggregating distributions to which many random processes formed of interacting components converge.

The key insight which explains this, originating from the maximum entropy principle and developed to a high level by \citet{frank2009common} is that the reason these distributions arise, and the reason they are what they are, is due to a maximum entropy process. If we consider some noisy process of interaction, which maintains only certain invariant properties of the distribution, then as more processes are included, or aggregated, then the overall distribution tends inexorably towards the maximum entropy distribution which satisfies those invariants as constraints. If the only information preserved through interactions is about variance, then a Gaussian distribution results. This maximum entropy approach generalizes the central limit theorem, which only considers the case of the addition, or convolution, of many random variables, since the maximum entropy approach demonstrates that \emph{any process of aggregation that preserves only variance} will inexorably tend towards the Gaussian. Similarly, \citet{frank2009common, frank2016common} extended this argument to show that many other well known distributions in classical statistics can be derived directly from maximum entropy principles given different constraints. For instance,  binomial distributions arise as the maximum entropy solution when we only care about numbers of a certain kind of event and not their order, and power-law distributions arise when the aggregation process preserves the geometric mean.

At the lowest level, the invariant quantities preserved through interaction are often those invariants relating (by Noether's theorem) to symmetries in the underlying physical laws (\citep{hanc2004symmetries,blanchard2012variational,lanczos2012variational}). For instance, the Boltzmann distribution in statistical physics derives entirely from the fact that the \emph{energy} of a (closed) physical system is conserved no matter the interactions between its components, and this energy conservation rule derives entirely from the time-translation symmetry of space-time, and we should expect other regular distributions to arise in other circumstances due to the conservation of other quantities, such as angular momentum or charge. These natural invariants mean that, at the very least, we should expect to be able to understand physical systems composed of many interacting variables based upon the low-dimensional summaries comprising the parameters of the maximum entropy distributions such systems naturally form, given the constraints imposed by these invariants. These parameters then become our first abstractions, which often allow us to almost perfectly model physical systems comprising of many trillions of interacting elements using only a few real numbers. 

We argue that this process of aggregation giving rise to well-defined and mathematically simple maximum entropy distributions with just a few key parameters, is ultimately what often leads to the existence of good abstractions to define a system. Specifically, since the maximum entropy distribution can be described in a few parameters, and many interactions involving that system only depend on the shape of the distribution as a whole, not the specifics of any event making up the distribution, then these interactions can also be predicted primarily on the basis of the maximum entropy parameters. This idea can then be extended hierarchically as well, where the maximum entropy distributions themselves become the interacting elements of a higher level process, which itself converges to another maximum entropy distribution which can then be described with just a few higher-level parameters. This hierarchical aggregation of maximum entropy distributions may allow for more complex `composite' constraints and hence distributions to be built out of simpler constraints while maintaining the ability to be described by (relatively) few parameters relative to the number of `base states' of the system. While the basic levels of aggregation in terms of preserving simple information such as means, variances, and ratios have been well worked out, as reviewed very clearly here \citep{frank2009common,frank2016common}, whether such composite distributions, which maintain more complex information about their underlying distributions, and thus have a more complex, but still finitely describable maximum entropy form, leading to more complex abstractions remains unknown and must be worked out precisely in future work. Nevertheless, while this is simply an intuitive sketch of an argument, we believe it provides a strong case for why we should naturally expect aggregation of many independent processes under some informational constraints which arise from the specifics of the interaction process to lead to limiting aggregate distributions which are easily describable with just a few parameters, and then these parameters can ultimately serve as our abstractions which can achieve high predictive accuracy of the system and its behaviour while containing vastly less information than the sum of the microstates of the system.

Since the notion of aggregating or limiting maximum entropy definitions can provide us with an understanding of why and how abstractable systems may come into being, it also simultaneously provides us with a set of conditions for when we might expect abstractable systems to arise: namely, those where we expect aggregating distributions to arise, which are those systems composed of many interacting random variables, which interact in a way which preserves only specific types of information. These conditions necessary for aggregation appear to arise regularly in the real world for a number of reasons. 

The key necessary quantities are fundamentally uncertainty and distance, either spatially or temporally or both. Uncertainty, which may relate to epistemic uncertainty of the observer or else intrinsic stochasticity in the dynamics (or both) is necessary to ensure that low level information is slowly wiped out over time, meaning that only high level informational constraints are preserved over multiple interactions. This property is necessary for all aggregating distributions -- for instance in the central limit theorem -- to ensure that noise from many different variables interferes with and destroys the information contained within other variables during their interactions. It is important to note here that epistemic uncertainty plays exactly the same role as noise here, even if there is no intrinsic stochasticity. If there is any errors and unknowns in the initial specification of a system with many interacting parts, of the kind that are described by limiting distributions, then interactions between components will tend to distribute and increase our epistemic uncertainty about the system until the uncertainty tends to the maximum possible (the maximum entropy distribution). This idea is identical to that of chaos in dynamical systems theory, whereby small uncertainties in the initial state of the system propagate exponentially until the microscopic state of the system rapidly becomes unpredictable, even with known, deterministic dynamics. Systems composed of extremely large numbers of interacting elements are almost always chaotic in their microscopic details \footnote{If the 3-body problem in classical mechanics is chaotic, imagine how chaotic the $6\times 10^{23}$-body problem must be}. It is this microscopic chaos, then, that ultimately allows for aggregating distributions to arise, and thus underpins the formation of predictable macrophenomena (or abstractions). The theory of maximum entropy thus provides a crucial link between reproducible and abstractable macrophenomena, and the microscopic chaos that underpins them. Conversely, microscopic chaos is what ultimately justifies the principle of maximum entropy. Namely, uncertainty in a chaotic system will increase exponentially until it hits the structural limit given by the information that is maintained across interactions, rendering the maximum entropy configuration the exponentially most likely solution \citep{jaynes1985macroscopic}.

The second necessary precondition for the observation of abstractions is distance, in terms of time or space. This is necessary to allow enough interactions amid the system elements for the uncertainty in those distributions to aggregate into a maximum entropy form. From a statistical perspective, this is because distance is what enables a large enough `sample size' to accumulate to enable the `data' to coalesce into a recognizable -- i.e. maximum entropy -- distribution. Heuristically, distance lets us accumulate the `large' in the law of large numbers. Things that are `far' from us look simpler, and can often be described accurately (in terms of their interactions with us) with fewer parameters. Things that are close to us appear more complex and their interactions with us are much more nuanced and require more parameters to describe. From the perspective of a gas molecule in the eponymous box of gas, the average energy of the gas does not seem like a useful description of the behaviour it sees around it. Conversely, from the perspective of a scientist wishing to explain its macroscopic properties, things like the average energy are perfectly sufficient. And, from the perspective of an alien in a far-off galaxy, any properties of the box of gas are irrelevant (and likely unknowable). All it needs to describe our interactions with it are extremely coarse-grained properties like the mass, centre of mass, and average luminance of the Milky Way galaxy. In some sense this is an obvious truism, but an important one. If we imagine all interactions between everything as taking place on some giant causal graph, then the distance between any two systems can be computed as the path length between them. At every node along this path, uncertainty or intrinsic noise propagates interferes with the original signal and reduces the information which can be retrieved at the end of the path, with the exception of potentially preserving some aggregate information. Over the course of this path, more and more interacting systems are aggregated together until the entropy increases to its structural limit, and the signal can be entirely represented as a maximum entropy distribution with just a few parameters. Then, if the path is long enough, this process continues recursively, with more information being lost, and the original maximum entropy distribution becomes merely an interacting element of a higher-order maximum entropy distribution, and so on, until entire swathes of the graph can be described relatively well purely in terms of a few abstraction parameters.

\subsection{Are there `natural' abstractions?}

Thus far, we have been considering abstractions considering only a single arbitrary query set, but this story of abstractions deriving from maximum entropy solutions arising due to aggregation suggests that there may be a set of `natural' abstractions for a given system -- namely the parameters of the maximum entropy distribution which describes the system. These parameters, or the natural abstraction of the system, should be able to perfectly answer any query which only needs information about the system's distribution to answer, rather than information about a particular element of the underlying system. Similarly, if maximum entropy distributions of systems can interact to produce more complex maximum entropy distributions with a more complex set of interaction constraints, as postulated previously, then the parameters which parametrize these constraints will become a higher level set of abstractions.

It may also be possible to generalize this idea, and ask, for any arbitrary system distribution, whether there exists some set of natural abstractions which can match any query concerning distributional properties of the system. Specifically, we wish to derive a measure of the universal `abstractability' of a distribution which is the amount of information in an abstraction needed to specify the full distribution of the system.  This question corresponds to whether a given arbitrary distribution can be compressed and represented as the maximum entropy solution to a set of finite constraints, and is closely related to the question of the extent to which an arbitrary distribution can be compressed, or represented as a program, and thus has close relations to Kolmogorov Complexity \citep{kolmogorov1965three,chaitin1977algorithmic} and algorithmic information theory.

We now consider how we might formalize this notion of a universal kind of `abstractability'. We start with some distribution $p(x)$, and we wish to measure how easily it can be encoded to a given degree of accuracy. The key problem is to develop a method to encode an arbitrary distribution varying over a continuum. Here, we assume that we can encode this arbitrary distribution as an infinite mixture of tractable exponential-family distributions. Here, without loss of generality, we assume that these exponential mixture distributions are Gaussian.
\begin{align*}
    p(x) \approx \frac{1}{Z} \sum_{i=0}^\infty \pi(x_i) \mathcal{N}(x; \mu_i, \sigma_i) \numberthis
\end{align*}
where $\pi(x_i)$ is the probability that weights that particular mixture density, $\mu_i, \sigma_i$ are the parameters of the mixture Gaussian, and $Z$ is an arbitrary normalizing constant that may be necessary. We know that this infinite mixture density can approximate any arbitrary distribution, since the value of the real distribution $p(x)$ at a point can simply be approximated by an infinitely narrow Gaussian distribution $\sigma \rightarrow 0$ with a mean of that point $\mu_i = x_i$, and the mixture weight equal to the probability $\pi(x_i) = p(x_i)$. The constraint that the mixture density coefficients must sum to one is not problematic since the real distribution shares the same constraint. While, we know we can therefore \emph{always} exactly approximate any distribution with an infinite mixture, there may be many other ways to approximate it. If there are, then we can derive a measure of the universal abstractbility of the distribution as the \emph{entropy of the mixture distribution} $\pi(x)$. If we are to use $\pi(x)$ as a measure of abstractability, then to make it a meaningful measure of the true abstractability, we must choose the minimum entropy solution of $\pi(x)$ under the constraint that the full set of $\pi(x), \mu_i, \sigma_i$ equals the real distribution. If we formalize this mathematically using Lagrange multipliers, we define the universal abstractability measure $\mathcal{A}[p(x)]$ as follows,
\begin{align*}
\label{abstractability_def}
    \mathcal{A}[p(x)] &= \mathbf{H}[\pi^*(x)] \\
    \pi^*(x) &= \underset{\pi(x),\mu_i,\sigma_i}{argmin} \, \mathbf{H}[\pi(x)] + \lambda \big(\frac{1}{Z} \sum_{i=0}^\infty \pi(x_i) \mathcal{N}(x; \mu_i, \sigma_i) - p(x) \big) \numberthis
\end{align*}

Let's consider the extreme cases. Suppose that the distribution $p(x)$ simply is an exponential-family (Gaussian) distributions $p(x) = \mathcal{N}(x; \bar{\mu}, \bar{\sigma})$. In this case, the minimum entropy mixture distribution is to set one mixture coefficient to 1 and the rest 0, and then set the parameters of that distribution to the parameters of the real distribution $\mu_i = \bar{\mu}; \sigma_i = \bar{\sigma}$. In this case, the abstractability $\mathcal{A} = \mathbf{H}[\pi^*(x)] = \mathbf{H}[\delta(x - x_i)] = 0$ ($\mathcal{A}$ really represents `un-abstractability' so that the smallest value it can take is $0$, meaning that all the information in the real distribution $p(x)$ can be abstracted). Conversely, suppose that we cannot abstract anything and are forced to match every point of $p(x)$ with its own mixture distribution. In this case, we have that $\mathcal{A} = \mathbf{H}[\pi(x)] = \mathbf{H}[p(x)]$, or that the worst-case abstractability is simply the entropy, or Shannon information content of the distribution \footnote{While it may be possible to have solutions to the mixture density problem with higher mixture entropies than the actual distribution, for instance a uniform $\pi(x)$, these possibilities are eliminated by the minimum entropy constraint in Eq \ref{abstractability_def}, since we know the $\pi(x) = p(x)$ solution always exists, it provides a strict upper bound on the abstractability}. Perhaps unsurprisingly, this highlights the close connection notions of abstractability have with information theory and Kolmogorov Complexity and Minimum Description Length, where we consider the concept of `universal abstractability' to be the same as its compressibility and closely related to its Kolmogorov complexity (i.e. the length of the smallest program needed to generate the object. This approach here effectively generalizes these ideas from discrete strings to continuous distributions. Importantly, the universal abstractability of a distribution is distinct from just the Shannon entropy of the distribution (the entropy provides an upper bound), as can be straightforwardly seen in the case where the distribution $p(x)$ takes the form of a single exponential family distribution. The precise relationship between the measure of universal abstractability proposed here and Kolmogorov complexity remains to be elaborated upon in future work. There is also a close connection between our conception of density approximation using an infinite mixture distribution and the field of nonparametric statistics which follow a similar approach of trying to model the data distribution using an implicitly infinite mixture distribution \citep{wasserman2006all,higgins2004introduction}. A key difference to our proposed method is the way the mixture distribution is computed. Nonparametric methods typically define an explicit generative process which generates the mixture distribution $\pi(x)$ such as, for instance, the Dirichlet stick breaking process \citep{teh2010dirichlet}, or the Indian Buffet process \citep{griffiths2011indian}. By contrast, we find the mixture distribution through the solution of a variational maximum entropy problem (Eq \ref{abstractability_def}). Beyond this, the relationship between our method of estimating abstractability and nonparametric statistics remains to be fully elucidated.

\section{Discussion}

In this paper, we have endeavoured to lay out the potential beginnings for a mathematical theory of abstraction; a concept which is widely used and understood intuitively, and with close connections to folk notions of understanding, learning, and inference, but which has so far mostly eluded mathematical formalization and scrutiny. We have argued that the two key defining properties of an abstraction are that it is query-dependent, in that any given abstraction is only designed to answer certain questions about a system accurately, and that the abstraction throws away information which is not relevant to answering those queries. Secondly, we formalize this notion of `throwing away' information through the use of a maximum entropy system which represents the knowledge of the system given only by the abstraction itself, and we can use this definition to define an objective which intuitively corresponds to the `leakiness' of an abstraction as a basis for both evaluating the value of any specific abstraction against a query set, as well as potentially as an objective function which, when combined with a parameterised abstraction model could be used to learn abstractions from data.

While our definition of abstraction is closely related to classical ideas in statistics, it actually comes from a subtly different perspective. Statistics is ultimately concerned with modelling the data -- we infer parameters which can best explain the observed data. Maximum likelihood methods make this explicit, but this philosophy underpins other approaches as well. We, on the other hand, are not, in general, concerned with modelling the data itself, but only the answers to a set of queries on the data, which may or may not require all the information contained within the dataset to answer. We thus have a notion of relevant or irrelevant information which classical statistical methods do not, as shown by the close connection discussed earlier (and in Appendix C) between maximum likelihood and information-maximization. Indeed, on a mathematical level, our definitions can simply be considered to be the application of standard statistical methods and objectives applied to a set of functions of the data, rather than the data itself. Conversely, we can consider classical statistics to be concerned with learning abstractions which are specifically just coarse-grainings of the dataset, in that they allow for the most efficient possible reconstruction of the full dataset. A similar story plays out with the relationship between our notion of abstraction and other notions of complexity such as Kolmogorov' Algorithmic Complexity, and the ideas of Minimum Description Length coding \citep{grunwald2007minimum}. These methods are fundamentally interested in the information necessary to encode the data (or object) perfectly, or if there is a loss, ensuring that it is small and relatively uniformly distributed across the full data-space. Our notion of abstraction, on the other hand, does not try to preserve all the information in the data, but only keeps that which is necessary to answer queries from the query set and discards everything else. 

Our results have deep implications for our understanding of inferential and cognitive systems in general. Firstly, it may be the case that we should not necessarily model such systems through the lens of classical statistics where we aim only to reconstruct data that we are given, but instead through the lens of first understanding the kinds of queries which the system needs abstractions and representations to answer, and then the kinds of representations which can best answer those queries. This is of especial interest for evolved biological systems which have evolved not, in general, to perfectly reconstruct the world, but only to respond adaptively to the world. This means that we might expect that the abstractions or representations learnt by such cognitive systems should be heavily tuned for the type of queries that are often necessary for survival and reproduction rather trying to build a detailed veridicial model of all the detail in the world. This view has close links to enactivist and embodied ideas in the philosophy of cognition which hold, at a broad level, that the key purpose of cognition is in enabling adaptive action rather than simply trying to understand the world in an unbiased way. Cognitive agents built according to such ideas might have representations and models of the world which are substantially `wrong' from the viewpoint of accurately representing the world, but which nevertheless are able to drive adaptive and successful action in the circumstances such agents usually find themselves. Our definition of abstraction can be thought of as providing a comprehensive mathematical formalism which can make these ideas precise.


Finally, it is important to note that the framework here is only preliminary and may be subject to change. Specifically, while we have attempted to put together a consistent and precise theory of abstraction, it is still undetermined whether this is actually the optimal or most useful definition of abstraction available. We believe that our definition matches many of the properties of our intuitive notion of abstraction, while also making them precise and available in a computationally tractable framework which permits the learning and inference of abstractions from both static datasets and dynamical systems. Nevertheless, it remains an open question whether the definitions of abstraction given here are the most felicitous possible, and in general the goal efficiently learning the optimal abstractions from data is still a long way away.

A key hurdle the theory must clear before it becomes useful is whether it is computable in practice and empirically leads to good results on learning benchmarks, especially if the optimization of Equation \ref{Abstraction_objective} actually leads to learning useful or recognizable abstractions given ecologically valid query sets. The computation of the maximum entropy system seems likely to be the key bottleneck in many cases, although for simple cases it can be worked out analytically using standard constrained optimization methods. Developing methods to make computing and optimizing Eq \ref{Abstraction_objective} and Eq \ref{maxent_variational_problem} efficiently, as well as designing statistical methods containing the right inductive biases to make abstraction learning straightforward remains an open and challenging task. With regards to applications for machine learning, one clear suggestion by our theory is simply that practitioners think much more deeply about the \emph{kinds of query} they actually want to make of a given system or dataset, and then train their system against those queries, and especially consider training against a large number of queries if that is desired, instead of just optimizing one proxy. For machine learning, this would consist of training with \emph{many loss functions} simultaneously, an area which is largely unexplored, and may yet yield substantial improvements to performance, robustness, and generalizability for a relatively small effort in understanding exactly what queries are desirable, and then designing an objective function that optimizes against those queries.

\section{Acknowledgements}
I wish to thank the many discussions with and edits by Christopher L Buckley, which substantially improved this work and especially the mathematical notation, and the contribution of Mycah Banks in helping proofread the document and prepare the figures. My work here owes a substantial intellectual debt to John Wentsworth whose Lesswrong articles on abstraction provided many useful insights and got me interested in providing a mathematical formulation of abstraction in the first place. The section on why abstractions are possible is heavily intellectually indebted to S.A Frank's papers on aggregation and limiting processes to a much greater degree than can be established by a simple citation. 

\bibliography{cites.bib}

\section{Appendix A: Derivation of the Gaussian Distribution from Maximum Entropy Principles}

Here, we derive the Gaussian distribution directly from the maximum entropy principle where we aim to maximize the entropy distribution based only on knowing a fixed variance of the distribution $\int dx \, p(x)(x - \mu)^2 = \sigma^2$. The first step is to write out the maximum entropy objective.
\begin{align*}
    \mathcal{L} = -\int dx \, p(x) \ln p(x) - \lambda_0 \big( \int dx \, p(x) - 1 \big) - \lambda_1 \big( \int dx \, p(x) (x - \mu)^2 - \sigma^2 \big) \numberthis
\end{align*}
For which we have the solution for $p(x)$ as,
\begin{align*}
\label{gaussian_pstar}
    \frac{\partial \mathcal{L}}{\partial p(x)} = 0 \implies p^*(x) = e^{-(\lambda_0 - 1)}e^{- \lambda_1 (x - \mu)^2} \numberthis
\end{align*}
Then, we substitute this into the first (normalization) constraint to get,
\begin{align*}
\label{gaussian_constraint_1}
    \int dx \, e^{-(\lambda_0 - 1)}e^{- \lambda_1 (x - \mu)^2} &= 1 \\
    e^{-(\lambda_0 - 1)} \int dx \, e^{- \lambda_1 (x - \mu)^2} &= 1 \\
    e^{-(\lambda_0 - 1)} \int dz \, e^{- \lambda_1 z^2} &= 1 \\
    e^{-(\lambda_0 - 1)} \sqrt{\frac{\pi}{\lambda_1}} &= 1 \\
     \sqrt{\frac{\pi}{\lambda_1}} &= e^{\lambda_0 - 1} \numberthis
\end{align*}
Where we have undertaken a change of variables $z = (x - \mu)$ and applied the Gaussian integral $\int dx \, e^{-x^2} = \sqrt{\pi}$. Given this relation, we substitute into the second (variance) condition,
\begin{align*}
\label{gaussian_constraint_2}
    e^{-(\lambda_0 - 1)} \int dx \, e^{\lambda_1 (x - \mu)^2} (x-\mu)^2 &= \sigma^2 \\
    e^{-(\lambda_0 - 1)} \int dz \, e^{\lambda_1z^2} z^2 &= \sigma^2 \\
    e^{-(\lambda_0 - 1)} \frac{1}{2}\sqrt{\frac{\pi}{\lambda_1^3}} &= \sigma^2 \\
    e^{(\lambda_0 - 1)} \sigma^2 &= \frac{1}{2\lambda_1} \sqrt{\frac{\pi}{\lambda_1}} \numberthis
\end{align*}

Then, putting together Eq \ref{gaussian_constraint_1} and Eq \ref{gaussian_constraint_2}, we have that,
\begin{align*}
    e^{\lambda_0 - 1} &= \sqrt{\frac{\pi}{\lambda_1}} = 2 \sigma^2 \lambda_1 e^{(\lambda_0 - 1)} \\
    &\implies 2\sigma^2 \lambda_1 = 1 \\
    &\implies \lambda_1 = \frac{1}{2 \sigma^2} \numberthis
\end{align*}
Now that we have $\lambda_1$, we can substitute back to get $\lambda_0$,
\begin{align*}
    e^{\lambda_0 - 1} &= \sqrt{\frac{\pi}{\lambda_1}} \\
    &= \sqrt{2 \pi \sigma^2} \\
    &\implies \lambda_0 -1 = \ln  \sqrt{2 \pi \sigma^2} \numberthis
\end{align*}
Substituting our expressions for the Lagrange multipliers $\lambda_0$ and $\lambda_1$ back into our expression for $p^*(x)$ (Eq \ref{gaussian_pstar}) gives us,
\begin{align*}
    p^*(x) &= e^{-(\lambda_0 - 1)}e^{- \lambda_1 (x - \mu)^2} \\
    &= e^{-\ln  \sqrt{2 \pi \sigma^2} - \frac{1}{2 \sigma^2}(x -\mu)^2} \\
    &= \frac{1}{\sqrt{2 \pi} \sigma}e^{-\frac{(x-\mu)^2}{2 \sigma^2}} \numberthis
\end{align*}
which is the Gaussian pdf.

\section{Appendix B: Relationship of the Maximum Entropy Principle to Variational Inference and Variational Free Energy Minimization}

The maximum entropy principle shares a close and interesting relationship with variational inference, which is a method to approximate optimal Bayesian inference, which involves the minimization of a variational free energy functional \citep{beal2003variational,blei2017variational,hinton1994autoencoders,wainwright2008graphical}. Indeed, variational inference originated in statistical physics and involves an information-theoretic analogue of an unimpeachable thermodynamic quantity -- the thermodynamic free energy. Here we explore connections between these two theories, and ultimately show that variational inference is a subtle generalization of the maximum entropy principle to the case where the energy term (which represents the constraints in the maximum entropy formulation) are not fixed but depend on the state of the system, and can thus be maximized in parallel with the entropy. The maximum entropy principle, by contrast, treats the energy term as a fixed constraint which is not affected by the system state, and thus turns the optimization of the variational free energy from an unconstrained to a constrained optimization problem.


The maximum entropy principle proposes that given certain knowledge about macroscopic properties of a system -- such that its average energy $\bar{E} = \langle E \rangle = \sum_i p(x_i) E_i$ is equal to some constant, we can derive the most likely probability distribution over the states as the one that maximizes their entropy while fulfilling the constraint on the average entropy,
\begin{align*}
    p^*(x_i) = \underset{p(x_i)}{argmax} \, \mathbf{H}[p(x_i)] - \sum_i \lambda_i \mathcal{E}_i \numberthis
\end{align*}
where $\mathcal{E}$ represents the set of constraints -- for instance that the average energy equals a known constant $\sum_i p(x_i) E_i = \bar{E}$, and that the probability distribution of the states sums to one $\sum_i p(x_i) = 1$, and $\lambda$ represents the lagrange multiplier. More generally, we can write this as minimizing a free-energy objective which is simply the energy minus the entropy,
\begin{align*}
    p^*(x_i) &= \underset{p(x_i)}{argmin} \, \mathcal{F} \\
    \mathcal{F} &= \underbrace{\mathcal{E}}_{\text{Energy}} - \underbrace{\mathbf{H}}_{\text{Entropy}} \numberthis
\end{align*}
This functional is known as the free energy of a system in thermodynamics and is central to the field. 
Conversely, in Bayesian inference, we often want to infer posterior distributions of some latent variables $x$ given some data $o$. By Bayes' rule we can write $p(x | o) = \frac{p(o,x)}{p(o)}$ where $p(o) = \int dx p(x,o)$ is known as the marginal likelihood and is often intractable to compute due to the integration over all latent variables required. If the marginal likelihood is intractable, then the Bayesian posterior will also be intractable as the marginal likelihood is precisely the normalization condition for the posterior. While many approaches have been developed in the literature for approximating the true Bayesian posteriors, one approach is known as variational inference, which instead postulates a separate `approximate posterior' distribution $q(x | o)$ which is designed to be tractable, and to optimize this approximate posterior by minimizing the divergence between it and the true posterior \citep{beal2003variational,wainwright2008graphical},
\begin{align*}
    q(x | o)^* = \underset{q(x | o)}{argmin} \, KL \big[ q(x | o) || p(x | o) \big]  \numberthis
\end{align*}
However, this divergence is itself intractable due to containing the intracatable posterior. Instead, we minimize an upper bound on this divergence known as the variational free energy $\mathcal{F}$. By minimizing this bound, the approximate posterior $q(x | o)$ converges to the true posterior.
\begin{align*}
    \mathcal{F} &= \mathbb{E}_{q(x | o)} \big[ \ln q(x | o) - \ln p(o,x) \big] \\
    &= KL \big[ q(x | o) || p(x | o) \big] - \ln p(o) \\
    &\geq  KL \big[ q(x | o) || p(x | o) \big] \numberthis
\end{align*}
Crucially, this `variational free energy' can be decomposed into precisely the energy and entropy terms of the free energy minimized by the maximum entropy principle,
\begin{align*}
    \mathcal{F} &= \mathbb{E}_{q(x | o)}[\ln q(x | o)] - \mathbb{E}_{q(x | o)}[\ln p(o,x)] \\
    &= -\mathbf{H} + \mathcal{E} \numberthis
\end{align*}
Where $ -\mathbf{H} = \mathbb{E}_{q(x | o)}[\ln q(x | o)]$ and $\mathcal{E} =  - \mathbb{E}_{q(x | o)}[\ln p(o,x)]$. So if they both involve minimizing a free energy functional, how do they differ? To understand the close relationship, we must first understand how to consider the maximum entropy objective as encoding an inference problem. First, it is clear that to phrase the constrained optimization as inference, we can define the values the constraints take to be our `observations' or `data' within the paradigm of Bayesian inference. This make sense, as the values of the constraints, such as the average energy of the box of gas, are precisely the `data' we work with when trying to infer the maximum entropy distribution of a system. Secondly, since the distribution $p(x)$ over the system states is what we are trying to infer, and which really represents our beliefs about the system states rather than the system state itself, it makes sense to simply consider the maximum entropy distribution which we solve for as our approximate variational posterior, which allows us to make the identification $p_{maxent}(x) = q(x | o)$. This identification allows us to identify the entropy term in the variational free energy with the entropy term in the maximum entropy objective. All that remains now is to understand how the energy terms of the two objectives relate. Crucially, we need to relate the constraints of the maximum entropy problem to the probabilistic generative model of variational inference. This can be done quite straightforwardly by assuming a factorized Gaussian form for the generative model with a set of independent constraints. That is, we define $p(o,x) = \prod_i p(o_i, x) = \prod_i \mathcal{N}(o_i; f(x_i), \sigma)$ where $f_i(x)$ encodes a specific constraint function -- i.e. for the average energy $f_i(x) = \sum_j p(x_j)E(x_j)$ and $o_i = c_i$ where $c_i$ is a constraint value -- i.e. the average energy equals some constant $c_i$. This allows us to write the energy term of the free energy as,
\begin{align*}
   \mathcal{E} =  \mathbb{E}_{q(x | o)}[\ln p(o,x)] &= \sum_i \mathbb{E}_{p(x)} \ln p(o_i, x) \\
   &= -\sum_i \mathbb{E}_{p(x)} \big[ \frac{1}{2 \pi \sigma^2}(c_i - f_i(x))^2 \big] + \ln C
\end{align*}

We can see this term approximates the constraint, since the best possible solution will be when $\mathcal{E} = 0$ or when the constraint is satisfied such that $c_i = f_i(x)$. This then allows us to identify the variance of this Gaussian with the lagrange multiplier by defining $\lambda = \frac{1}{2\pi\sigma^2}$. Thus, the constraint becomes tight as $\lambda \rightarrow \infty$ or, equivalently $\sigma \rightarrow 0$. This set of definitions allows us to translate maximum entropy problems into variational inference problems and vice versa, thus allowing us to see the close connections between these two approaches. Importantly, we can see that the maximum entropy problem can be construed as a variational inference problem with the approximate posterior distribution being the our maximum entropy belief distribution over the system state, and with this Gaussian generative models which considers the constraint values as Bayesian observations, and the constraint functions as the prior over the system states, and additionally allows for the identification of the variance of the generative model with the lagrange multiplier. This means that we can consider the maximum entropy problem to be a special case of variational inference under certain conditions, and may also lead to the ability to extend the maximum entropy approach to handle uncertainty natively, for instance we can now represent uncertainty in our constraint values in a natural manner.


\section{Appendix C: Relationship to the principle of Information-Maximization (InfoMax)}

A widely used principle in machine learning and neuroscience is the idea that to learn good latent representations we should try to \emph{maximize} their mutual information with the data. The intuition is that if the latent variables contain large amounts of mutual information with the data, then they must presumably be good representations for it. Here we show how this principle relates to our definition of abstraction. Specifically, we show that this objective emerges naturally from simple maximum likelihood maximization under the assumption that the abstraction or statistical parameters is a deterministic function of the data, and thus mutual information maximization is an inseparable consequence of maximum likelihood estimation. Secondly, we show that this fact also naturally extends to our notion of abstraction, whereby optimizing our abstraction objective naturally tends to maximize the mutual information between the query and the abstraction distribution -- thus ensuring that the abstraction vector stores as much information as possible about the query (but not the data) as possible. Finally, in the presence of latent variables, we show how this mutual information maximization objective results in the minimization the conditional mutual information between the latent variables and the data, given the deterministic parameters, reproducing classical results in coding and information theory showing that non-redundant (seemingly random) codes provide the greatest possible information, as well as their application into neuroscience by Barlow's principle of minimum redundancy \citep{barlow1961coding}.

We begin with our reconstruction abstraction objective, which we have shown reduces to a maximum likelihood objective.
\begin{align*}
    a^* &= \underset{a}{argmin} \, \mathbb{E}_{p(D)}\big[ \KL[p(x) || \tilde{p}(x | a)] \big] \\
    &= \underset{a}{argmin} \, \mathbb{E}_{p(D)}\big[ \ln p(x) - \ln \tilde{p}(x | a) \big] \numberthis
\end{align*}
Where the double expectation under $x$ disappears since the data expectation is an empirical distribution consisting solely of delta distributions around the observed data-points. Then, if we assume that the abstraction distribution is a deterministic function of the data, we can write this as,
\begin{align*}
    a^* &= \underset{a}{argmin} \, \mathbb{E}_{p(D)p(a | D)} \big[ \ln p(x) - \ln \tilde{p}(x | a) \big] \\
    &= \underset{a}{argmin} \, -\mathbb{E}_{p(D)p(a | D)} \big[\ln \tilde{p}(x | a) - \ln p(x) \big] \\
    &= \underset{a}{argmax} \, \KL \big[ p(D,a) || p(D) p(a) \big] \\
    &= \underset{a}{argmax} \, \mathcal{I}[D,a] \numberthis
\end{align*}
Where we see that the objective directly corresponds to maximizing the mutual information between the data distribution and the abstraction. This can be generalized straightforwardly to the case with an arbitrary query-set $q$ as follows,
\begin{align*}
    a^* &= \underset{a}{argmin} \, \mathbb{E}_{p(q)p(D)}\KL \big[ Q(x)|| Q(m(a)) \big] \\
    &= \underset{a}{argmin} \, \mathbb{E}_{p(q)p(D)p(a | D)}\big[ \ln Q(x) - \ln Q(m(a)) \big] \\
    &= \underset{a}{argmin} \, -\mathbb{E}_{p(q)p(D)p(a | D)}\big[\ln p(q | a) -  \ln p(q | x) \big] \\
    &= \underset{a}{argmin} \, -\mathbb{E}_{p(q,a,D)}\big[\ln p(q | a) -  \ln p(q | x) + \ln p(a | x) - \ln p(a |  x) \big] \\ 
    &= \underset{a}{argmax} \, \KL \big[ p(q,a | x) || p(q | x)p(a|x) \big] \\
    &= \mathcal{I}[q, a | x] \numberthis
\end{align*}
Where we see that in the case of an arbitrary query, where the abstraction is simply a deterministic function of the data, we can see that minimizing the abstraction objective is equivalent to maximizing the mutual information of the query and abstraction, given the data.

Finally, returning to the case of the reconstruction query, supposing we no longer have the abstraction being a direct function of the data, but instead we take it to be an abstraction distribution $p(a)$ which depends on a deterministic parameter $\phi$ which is itself a deterministic function of the data. In probabilistic terms, this means that we can consider the abstractions $a$ now as latent variables while the $\phi$ are the parameters of the generative model. Thus, if we wish to maximize the mutual information between the data and the parameters, this mutual information can then be decomposed into two terms -- the mutual information between the data and both the latent variables and the parameters, minus the mutual information between the latent variables and data, given the parameters. 
\begin{align*}
    \mathcal{I}[x; \phi] = \mathcal{I}[x; a, \phi] - \mathcal{I}[a; x | \phi] \numberthis
\end{align*}
Maximizing $\mathcal{I}[x; \phi]$ thus entails minimizing $\mathcal{I}[a; x | \phi]$ which is the mutual information between the abstraction latent variables and the data, given the parameters. Effectively, minimizing this term reduces the `redundancy' of the data encoding -- in that there is a minimum of information about the data left in the latent variables; all information has been taken over by the parameters. This makes sense in terms of coding -- an ideal code would be highly non-redundant for which there is no predictable information between the latents and the data which is encoded in the latents since, by the tenets of information theory, that would mean a more efficient code could exist. This derivation thus ties the maximization of mutual information via maximum likelihood, and its generalization to arbitrary queries and abstraction to classical results in information theory and coding theory \citep{shannon1948mathematical,hamming1986coding}, as well as its extensions such as the minimum redundancy principle for neural systems \citep{barlow1961coding,bell1995information}.

\section{Appendix D: Applications to Machine Learning}

The immediate applications of our method to modern machine learning methods arise directly from the relationship of our notion of abstraction to classical statistical (maximum likelihood) learning discussed previously. Specifically, most machine learning systems can be considered as performing some kind of maximum likelihood inference, or related Bayesian or variational inference schemes. These can, in our framework, be interpreted as trying to learn abstractions under the reconstruction query $Q(x) = p(x)$. The key suggestion our framework would make to the field of machine learning is a conceptual one: consider whether the reconstruction query is \emph{really} the only query you are interested in learning about, whether you want abstractions that only try to reconstruct the data, or rather that can use other queries about the dataset. If the latter, then our method provides an immediate approach to do so.

Our abstraction objective (Eq \ref{Abstraction_objective}) is straightforwardly computable and can thus be optimized against the parameters of the abstraction function using automatic differentiation and gradient descent. The key additional idea arises in Eq \ref{query_average_obj} where we argue the true loss function to optimize against a query set is the average over all queries. This corresponds to a weighted sum of query functions, and thus in machine learning terminology, suggests that we should train machine learning models using a \emph{weighted sum of many loss functions simultaneously}, where each loss function corresponds to a specific query. This approach of optimizing using a large number of loss functions simultaneously has not been widely explored in the literature and thus may have the potential to yield significant improvements with relatively little cost.

Supporting evidence comes from the notion of regularization, which often adds additional penalty terms of various sorts to the loss function to discourage pathological behaviour (typically overfitting) of the optimizer when trained only upon a single loss function. This has been found to be highly beneficial in practice, and often careful regularisation and tuning is necessary to train many complex architectures successfully. From our perspective of abstraction, we argue that the key issue, which regularisation methods try to ameliorate, is that the neural networks are typically optimized only against a single loss function, or a single query. Conversely, we would argue that the real solution would be to figure out the precise query set that the network should be able to answer with its learned abstractions, and then optimize against the weighted sum of all the corresponding queries according to Equation \ref{query_average_obj}. We can think of this as a sort of very high dimensional multi-objective optimization problem, which generalizes standard regularisation methods, and may yield substantially more robust and effective abstractions than are learnt by just optimizing a single loss function.

\end{document}